%% file: sae_iclr.tex
\documentclass{article} 
\usepackage{iclr2025_conference,times,graphicx,booktabs}

\input{math_commands.tex}

\usepackage{hyperref}
\usepackage{url}
\usepackage{graphicx}
\usepackage{subcaption}

\title{Towards Interpretable and Inference-Optimal CoT Reasoning with Sparse Autoencoder-Guided Generation}

\author{Daniel Zhao\thanks{Equal Contribution}, \; Abhilash Shankarampeta$^*$, \; Lanxiang Hu$^*$, \; Tajana Rosing, \; Hao Zhang \\ University of California, San Diego \\ \texttt{\{djzhao, ashankarampeta, lah003, tajana, haozhang\}@ucsd.edu}}
\iclrfinalcopy 
\begin{document}

\maketitle

\input{text/0_abstract}
\input{text/1_intro}
\input{text/2_related}

\input{text/3_method}
\input{text/4_results}
\input{text/5_conclusion}

\newpage
\bibliography{reference}
\bibliographystyle{iclr2025_conference}

\newpage
\input{text/6_appendix}

\end{document}

%% file: math_commands.tex
\usepackage{amsmath,amsfonts,bm}

\def\figref#1{figure~\ref{#1}}

\def\eqref#1{equation~\ref{#1}}

\def\1{\bm{1}}

\DeclareMathAlphabet{\mathsfit}{\encodingdefault}{\sfdefault}{m}{sl}
\SetMathAlphabet{\mathsfit}{bold}{\encodingdefault}{\sfdefault}{bx}{n}



%% file: text/0_abstract.tex
\begin{abstract} 

We propose a novel method that leverages sparse autoencoders (SAEs) and clustering techniques to analyze the internal token representations of large language models (LLMs) and guide generations in mathematical reasoning tasks. Our approach first trains an SAE to generate sparse vector representations for training tokens, then applies $k$-means clustering to construct a graph where vertices represent token clusters and weighted edges capture sequential token transitions. Using this graph, we define an edge-weight based reward function to quantify adherence to established reasoning traces, thereby identifying exploitative reasoning trajectories. Additionally, we measure generation diversity from clustering to assess the extent of exploration. Our findings indicate that balancing both exploitation and exploration is crucial for achieving high accuracy in mathematical reasoning tasks. During generation, the SAE can serve as a scalable reward model to guide generations, ensuring a balanced trade-off between exploitation and exploration. This prevents extreme behaviors in either direction, ultimately fostering a higher-quality reasoning process in LLMs.

\end{abstract}

%% file: text/1_intro.tex
\section{Introduction}

Recent advancements in language model training show new opportunities beyond the pre-training scaling law with model size, compute and dataset size~\citep{kaplan2020scaling}. Inference-time scaling becomes increasingly important as reinforcement learning (RL) is incorporated into training pipelines for industry-grade LLMs~\citep{guo2025deepseek, openai2024o3, openai2024learning, kimi2025k1.5}. With RL, the models learn to explore rewards and new trajectories for problem solving that have never appeared in pre-training corpus~\citep{openai2024learning}. LLMs trained with such inference-time search capability have demonstrated superior performance in reasoning benchmarks~\citep{snell2024scaling, wu2024inference}.


On the other hand, as the number of inference tokens scales, it becomes more and more challenging to track rewards for intermediate steps. Common approaches employed to track rewards include string matching~\citep{wei2022chain} and using reward models~\citep{ma2023step}, but they are not scalable as the search space becomes larger for harder tasks. Manual inspection is also unscalable, given the amount of work required.

As a result, there is a call for a mechanistic technique to supervise the generation process and assign appropriate rewards for intermediate search steps. An automated supervision pipeline for longCoT generation is useful not only for generation but also for improving RL training itself.

At inference time, recent research has shown that there is excessive redundancy in longCoT after training with long Chain-of-Thought (CoT) data obtained from an SFT checkpoint~\citep{wang2025thoughts}. As a result, a mechanistic technique that assigns rewards to guide model generation in more promising directions has great potential to save more tokens and improve generation efficiency.

In training, despite emerging techniques such as adding a length penalty to training loss or adding another stage of long2short RL training~\citep{kimi2025k1.5}, RL-based approaches remain costly. A mechanistic technique that ranks trajectories based on their balance between exploitation (shorter generations for promising directions) and exploration (longer generations with more reversions) during RL enables methods like prioritized experience replay~\citep{liu2023prioritized}, facilitating more efficient learning and potentially leading to more optimal generations during inference.

In this paper, we present an unsupervised learning technique that uses the sparse autoencoder (SAE), which has been shown to learn interpretable representations~\citep{gao2024scalingevaluatingsparseautoencoders} to generate clustering and construct a prior knowledge graph by learning from an existing data set that contains correct mathematical reasoning trajectories~\citep{numina_math_7b}. We first train an SAE to generate sparse vector representations of reasoning trajectories in the training data. Then, we apply $k$-means clustering to construct a reference graph, where vertices represent token clusters that capture conceptual information, and edges encode probabilistic transitions between them.

Using this graph, we establish a reference framework to evaluate how closely a reasoning trajectory aligns with existing trajectories in the training corpus by tracking the nodes and edges it traverses. We also investigate various metrics to quantify cluster diversity across different reasoning trajectories, capturing the extent of exploration. Our findings highlight a striking balance between exploitation and exploration, and considering multiple exploration metrics are important to achieving strong performance in mathematical reasoning tasks. This work demonstrates the potential of leveraging rich representations extracted from SAEs and employing an unsupervised learning approach as a highly scalable reward model to guide high-quality CoT generation.

%% file: text/2_related.tex
\section{Related Work}
\subsection{Sparse autoencoders}
Sparse autoencoders (SAEs) \citep{makhzani2013k} have emerged as a powerful tool for disentangling and interpreting internal representations in large language models (LLMs) \citep{cunningham2023sparse, bricken2023monosemanticity, elhage2022toy}. These models map high-dimensional hidden states into sparsely activated feature spaces, enabling more precise control and interpretation of model behavior.

Recent work has focused on improving SAE scalability and training stability. \cite{gao2024scalingevaluatingsparseautoencoders} introduced TopK activation functions to enforce sparsity constraints more effectively, demonstrating success even with large models like GPT-4. \cite{rajamanoharan2024improving} proposed gated sparse autoencoders, separating active latent selection from activation magnitude estimation, leading to improved reconstruction-sparsity trade-offs and enabling training of extremely wide autoencoders with up to 16 million latents.

In the context of LLM interpretability, \cite{cunningham2023sparse} showed that SAEs can decompose activations into interpretable features corresponding to specific concepts, while \cite{marks2024enhancing} extended this work to discover sparse feature circuits that reveal causal relationships within models. Building on these advances, recent work has focused on leveraging SAEs for controlled generation: \cite{yang2025lf} developed LF-Steering to address semantic inconsistency, and \cite{yin2024direct} introduced Feature-level Preference Optimization (FPO) to improve alignment efficiency through sparse feature manipulation.

\subsection{Chain-of-Thought Reasoning in LLMs}

Chain-of-thought (CoT) reasoning has emerged as a powerful approach for enhancing language models' reasoning capabilities. First introduced by \cite{wei2022chain}, CoT prompting demonstrated that large language models can break down complex problems into intermediate reasoning steps by augmenting few-shot exemplars with step-by-step rationales. This discovery sparked numerous innovations in the field - from zero-shot CoT prompting with simple instructions \citep{kojima2022large} to program-guided reasoning like Program-of-Thought \citep{chen2022program} and PAL \citep{gao2023pal}. Recent work has explored more sophisticated structures beyond simple chains, such as Tree-of-Thoughts \citep{yao2024tree} enabling exploration of multiple reasoning paths, Graph-of-Thoughts \citep{besta2024graph} incorporating cycles and verification steps, and Self-Consistency \citep{wang2022self} leveraging multiple reasoning trajectories. The field has also seen advances in making CoT more robust through verification and refinement \citep{madaan2024self}, knowledge enhancement \citep{wang2023knowledgpt}, and efficient reasoning techniques \citep{chu2023survey, bi2024forest}. Most recently, research has revealed that CoT capabilities may be inherent in pre-trained models even without explicit prompting \citep{wang2024chain}, suggesting these reasoning patterns emerge naturally during pre-training but require appropriate decoding strategies to surface.

Recent work has demonstrated the advantages of longer-form Chain-of-Thought reasoning through expanded context windows. Models like OpenAI's o1 \citep{jaech2024openai} and Kimi k1.5 \citep{MoonshotAI} show that scaling CoT to 128k tokens enables more comprehensive reasoning with improved performance on complex tasks. While these long-form approaches incorporate additional cognitive processes like planning and reflection, they face challenges of computational cost and potential "overthinking" \citep{chen2024not}. This has motivated research into distilling long-form reasoning into efficient "short-CoT" formats through techniques like preference optimization \citep{rafailov2024direct} and reinforcement learning-based generation \citep{ahmadian-etal-2024-back, luo2025o1}, balancing reasoning depth with practical constraints.

%% file: text/3_method.tex
\section{Method}

This work proposes a method that leverages SAE and clustering techniques to analyze token representations for reasoning tasks. In this section, we provide a detailed description of our approach.

\subsection{Preliminary}

\begin{figure}
    \centering
    \includegraphics[width=1\linewidth]{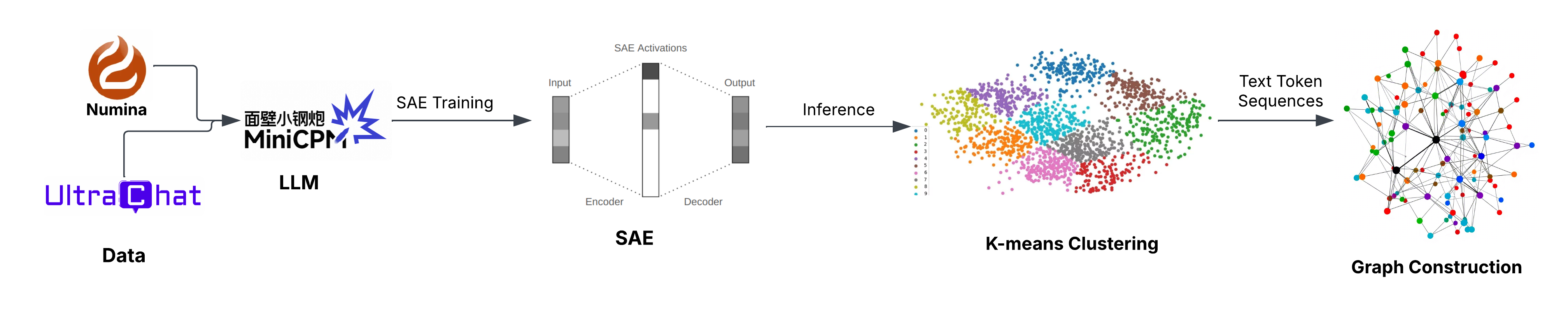}
    \caption{Flow of our pipeline}
    \label{fig:system-flow}
\end{figure}

\textbf{Sparse Autoencoder Training.} We first train a sparse autoencoder with a hidden dimension of $n = 8192$ à la \cite{gao2024scalingevaluatingsparseautoencoders}. Given an input token representation $\mathbf{x} \in \mathbb{R}^d$, the encoder maps $\mathbf{x}$ into a high-dimensional latent space as follows:
\begin{equation}
    \mathbf{z} = \mathrm{TopK}\left(W_{enc} \mathbf{x} + \mathbf{b}_{pre}\right),
\end{equation}
where $W_{enc} \in \mathbb{R}^{n \times d}$ is the encoder weight matrix, $\mathbf{b}_{pre} \in \mathbb{R}^{d}$ is the pre-bias, and $\mathrm{TopK}$ denotes an activation function that only keeps the k largest latents, zeroing the rest.

The decoder then reconstructs the input as:
\begin{equation}
    \hat{\mathbf{x}} = \left(W_{dec} \mathbf{z} + \mathbf{b}_{pre}\right),
\end{equation}
where $W_{dec} \in \mathbb{R}^{d \times n}$ and $\mathbf{b}_{pre} \in \mathbb{R}^{d}$ are the decoder weight matrix and pre-bias, respectively.

The training objective is simply to minimize the reconstruction loss:
\begin{equation}
    \mathcal{L}_{\text{rec}} = \| \mathbf{x} - \hat{\mathbf{x}} \|_2^2,
\end{equation}
The training data consists of a balanced mixture (50/50) of samples from the NuminaMath \citep{numina_math_7b} and Ultrachat \citep{ding2023enhancing} datasets, applied to three MiniCPM \citep{hu2024minicpmunveilingpotentialsmall} models (a 2B-parameter model, a 2B-parameter SFT model, and a 1B-parameter SFT model).

\textbf{Sparse Representation Extraction.} After training the SAE, inference is performed on 5,000 samples each from both the NuminaMath and Ultrachat datasets, corresponding to approximately 4 million tokens. For each token, the SAE outputs a list of the top activated latents together with their activation values. 

Let \( \mathcal{A}(\mathbf{x}) \subset \{1,2,\dots,8192\} \) denote the set of neuron indices returned by the SAE for token \( \mathbf{x} \), and let \( a_i \) denote the activation value corresponding to neuron \( i \in \mathcal{A}(\mathbf{x}) \). We then construct a sparse representation \( \tilde{h}(\mathbf{x}) \in \mathbb{R}^{8192} \) by placing each retrieved activation at its corresponding index and setting all other entries to zero:
\begin{equation}
    \label{eq:4}
    \tilde{h}(\mathbf{x})_i =
    \begin{cases}
        a_i, & \text{if } i \in \mathcal{A}(\mathbf{x}), \\
        0,   & \text{otherwise.}
    \end{cases}
\end{equation}
This sparse vector is subsequently used for downstream clustering and analysis.

\subsection{Clustering via \texorpdfstring{$k$}{k}-Means}

To group semantically related tokens, we apply the $k$-means clustering algorithm to a randomly selected subset of the sparse vectors $\tilde{h}(\mathbf{x})_i$ derived in \eqref{eq:4}. Let the set of cluster centroids be denoted by $\{\mathbf{c}_1, \mathbf{c}_2, \dots, \mathbf{c}_K\}$, where $K$ is the number of clusters. Each token's centroid is computed by minimizing the cosine distance between the token representations and the centroids. 

\begin{figure}
    \centering
    \includegraphics[width=0.95\linewidth]{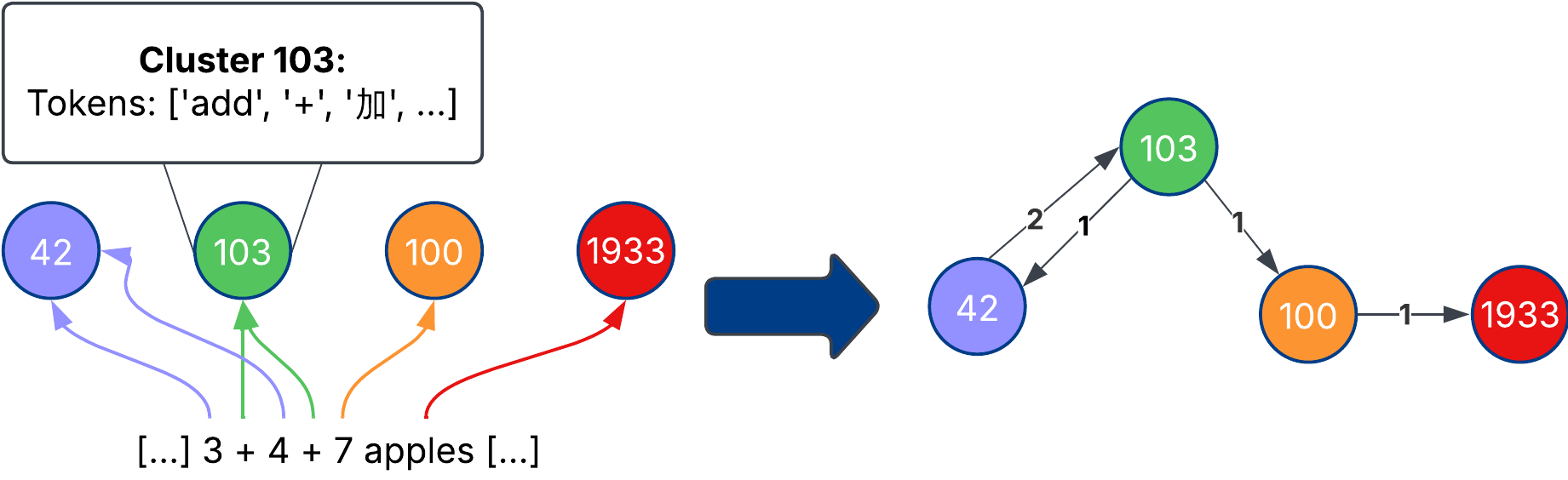}
    \caption{Sample construction of graph with the sentence ``3 + 4 + 7 apples". Left shows cluster assignment of tokens. Right shows conversion to edges and nodes.}
    \label{fig:graph-construction}
\end{figure}

\subsection{Graph Construction}

We construct a graph $\mathcal{G} = (\mathcal{V}, \mathcal{E})$, where each vertex $v \in \mathcal{V}$ represents a cluster (or node) and the edges capture sequential relationships between tokens. Importantly, this graph is constructed based \textbf{only on the tokens in the NuminaMath dataset}. Specifically, given a sequence of cluster assignments $\{c_1, c_2, \dots, c_N\}$ corresponding to the token sequence, the weight of the edge between clusters $i$ and $j$ is defined as:
\begin{equation}
    w_{i,j} = \sum_{t=1}^{N-1} \mathbb{I}\{c_t = i \land c_{t+1} = j\},
    \label{eq:6}
\end{equation}
where $\mathbb{I}\{\cdot\}$ is the indicator function that evaluates to 1 if its argument is true and 0 otherwise.

The resulting graph enables us to analyze the structural relationships between clusters via the edge weights. Since we build this graph only on the NuminaMath dataset, this analysis can provide insights into critical patterns and features that are relevant to reasoning and math-related tasks. 

We show an example of graph construction in \figref{fig:graph-construction}.

\subsection{Reward Model and the Explore-Exploit Trade-Off}

We use a simple reward model defined as follows to show that our clustering and graph construction algorithm is grounded. Let the reward $ R $ for a prompt $p$ be
\begin{equation}
    R(p) = \sum_{i=0}^{N-1} w_{p_i,p_{i+1}}
\end{equation}
where $N $ is the number of tokens in prompt $p$ and $p_i$ is the  $i^{\text{th}}$ token in $p$. The weight $w_{i,j}$ is computed as in \eqref{eq:6}. This reward function quantifies the extent to which a prompt adheres to the sequential patterns prevalent in the NuminaMath dataset.

\noindent \textbf{Justification of the Reward Model.} The reward model is justified by its interpretability and its direct connection to the structural relationships between token clusters. Since each edge weight $w_{i,j}$ captures the frequency of transitions between clusters $i$ and $j$, the cumulative reward $R(p)$ serves as an interpretable metric that reflects how well a given prompt aligns with established reasoning pathways; certain sequences of reasoning steps are more likely to lead to correct solutions.

\noindent \textbf{Explore vs. Exploit in the Graph.} In complex reasoning tasks, balancing exploration and exploitation is critical. In our framework, exploitation corresponds to following the highest weighted edges in the graph, which represent well-established and frequently observed token transitions. This type of greedy strategy reinforces conventional reasoning patterns that have proven successful in the NuminaMath dataset.

\begin{figure}
    \centering
    \includegraphics[width=1\linewidth]{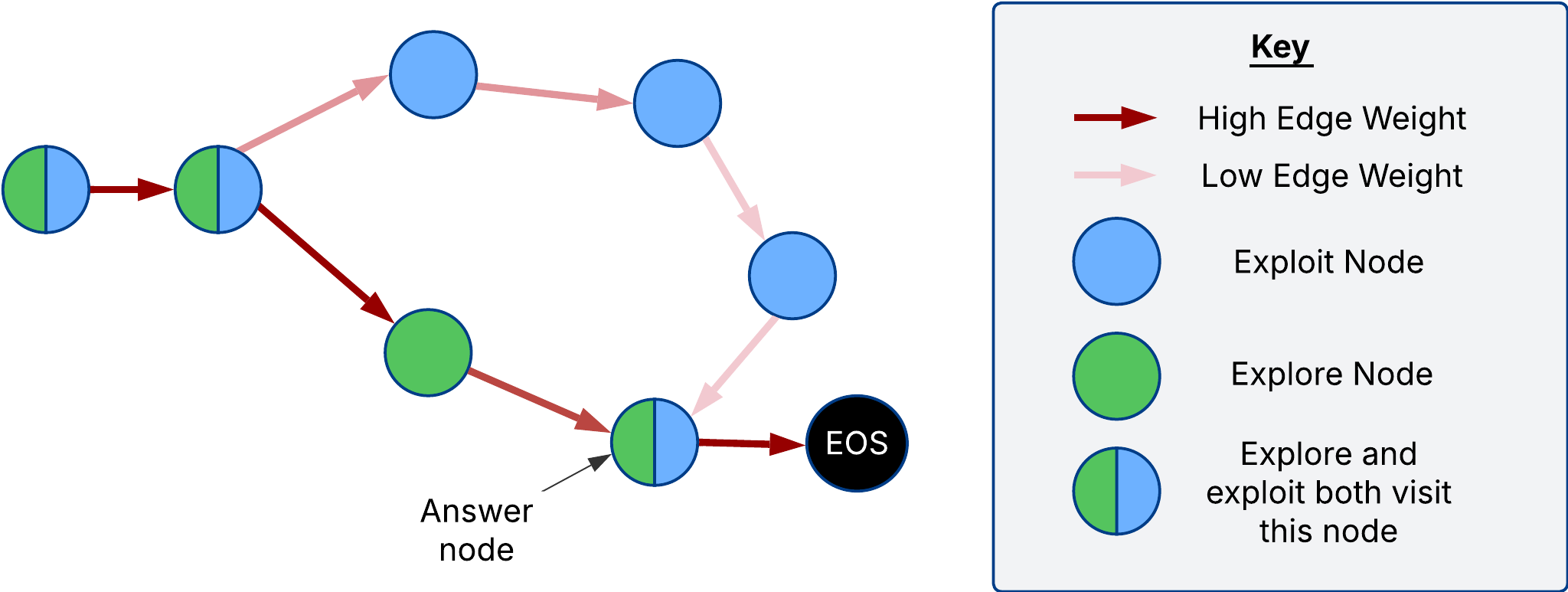}
    \caption{Graph showing example exploit vs. explore reasoning trajectories. Exploit takes the greediest approach, only transitioning through high weight edges. Explore takes lower weight edges but arrives at the same correct answer.}
    \label{fig:exploit-explore}
\end{figure}

Conversely, exploration involves deliberately probing transitions that are less frequent. Although these transitions may not be optimal in a strictly greedy sense, they can lead to the discovery of alternative reasoning pathways that might yield improved or more creative solutions. We illustrate this trade-off in \figref{fig:exploit-explore}.

\subsection{Metrics}

Let us define $G$ as the set of generated token sequences and $O$ as the set of original token sequences.

To assess the quality of the $G$ against $O$, we employ a combination of metrics that evaluate fidelity, diversity, and structural similarity. These metrics are designed to provide a comprehensive understanding of how well the generated sequences align with the original data in terms of content, variability, and structural patterns.

\subsubsection{Diversity Metrics}
\paragraph{Entropy.} Entropy is used to measure the diversity of cluster frequencies in the generated sequences. Higher entropy indicates greater diversity, which is desirable as it suggests that the generated sequences are not overly repetitive and exhibit a variability similar to that of the original sequences. The entropy of the generated sequences is compared to that of the original sequences to check the model's diversity in generation.

\subsubsection{Alignment Metrics}
\paragraph{Dynamic Time Warping (DTW).} DTW is employed to measure the alignment between sequences in $G$ and $O$. For each generated sequence, the minimum DTW distance from the original sequences is calculated using the SAE representation. The average of these distances provides a quantitative measure of structural similarity. Lower DTW distances indicate better alignment, suggesting that the generated sequences closely follow the structural patterns of the original data.


\paragraph{Distance Distribution.} The KL divergence is also used to compare distance distributions between consecutive tokens in the sequences. This metric evaluates whether the generated sequences maintain the same relational patterns between tokens as observed in the original sequences. Two approaches are used for this comparison: one using the direct representation of each element (SAE) and another using centroid-based representations obtained through k-means clustering.





%% file: text/4_results.tex
\section{Results \& Analysis}

\subsection{Graph-based Validation Metrics for Exploitation}

We compute the average reward per token for different datasets using the reward formulation described in \eqref{eq:6}) to evaluate how well our graph-based reward model captures established reasoning pathways. Table~\ref{tab:reward_results} summarizes the results for three models: MiniCPM-2B-128k, MiniCPM-2B-sft-bf16 and MiniCPM-1B-sft-bf16 \citep{hu2024minicpmunveilingpotentialsmall}.

\begin{table}[ht]
    \centering
    \begin{tabular}{lccc}
        \toprule
        \textbf{Dataset} & \textbf{MiniCPM-2B-128k} & \textbf{MiniCPM-2B-sft} & \textbf{MiniCPM-1B-sft} \\
        \midrule
        GSM8K       & 189.40  & 74.02  & 124.28 \\
        HumanEval   & 120.81  & 43.47  & 84.27  \\
        Wikipedia   & 93.32   & 5.27   & 11.43  \\
        NuminaMath  & 415.97  & 186.74 & 424.75 \\
        \bottomrule
    \end{tabular}
    \caption{Average per-token rewards computed using our graph-based reward model over 100 prompts per dataset. Higher rewards on reasoning-centric datasets (e.g., NuminaMath, GSM8K, HumanEval) versus lower rewards on non-reasoning data (Wikipedia) indicate stronger alignment with established reasoning patterns for each model.}
    \label{tab:reward_results}
\end{table}

Our results show that both models easily achieve the highest average rewards on the NuminaMath dataset; this should be given since we built our graph on this dataset.

However, we also see that reasoning patterns carry over to datasets like GSM8K \citep{cobbe2021gsm8k} and HumanEval \citep{chen2021evaluating}, with relatively high scores. Their contrast to the significantly lower rewards for the Wikipedia dataset (83.32, 5.27, 11.43), especially on the SFT models, demonstrates that the reward model is still sensitive to reasoning patterns across different types of data. Tokens that do not follow the prevalent reasoning pathways receive much lower rewards.

These results confirm that exploitation, i.e., greedily following the high-reward transitions encoded in the graph, can be a viable approach for guiding LLM generation in math and reasoning-related tasks. An exploitation-based approach may lead to more coherent and effective outputs. Overall, these metrics ground our graph construction method and show that higher rewards can help guide better generation.

\subsection{Comparative Analysis}

We evaluate the three MiniCPM language models on their ability to generate sequences that match the structural and distributional properties of the original data. The evaluation employs multiple metrics to assess accuracy, structural alignment (via DTW), distributional similarity (via KL divergence), and diversity (via entropy).

\subsubsection{Model Performance Overview}
The supervised fine-tuned models (MiniCPM-1B-sft and MiniCPM-2B-sft) demonstrate superior accuracy compared to the context-extended model (MiniCPM-2B-128k), as shown in Table \ref{tab:model-performance}. This comparison suggests that supervised fine-tuning has a more significant positive impact on sequence generation quality than extending the context length.
\begin{table}[h]
\centering
\begin{tabular}{lccc}
\toprule
\textbf{Model} & \textbf{Accuracy (\%)} & \textbf{Avg. Correct DTW} & \textbf{Avg. Incorrect DTW} \\
    \midrule
    MiniCPM-2B-128k & 39.87 & 17.05 & 20.23 \\
    MiniCPM-1B-sft & 52.00 & 4.66 & 5.55 \\
    MiniCPM-2B-sft & 53.40 & 31.35 & 36.87 \\
    \bottomrule
\end{tabular}
\caption{Comparison of model performance across sequence generation tasks, showing accuracy (in \%) and Dynamic Time Warping (DTW) distances for both correct and incorrect sequences. Lower DTW distances indicate better structural alignment with original sequences.}
\label{tab:model-performance}
\end{table}

\subsubsection{Structural Alignment Analysis}
Dynamic Time Warping (DTW) distances reveal distinct patterns across models. The MiniCPM-1B-sft model shows the lowest DTW distances, with correct sequences averaging 4.66 and incorrect sequences averaging 5.55, indicating superior structural alignment with the original sequences. In contrast, MiniCPM-2B-sft exhibits the highest DTW distances (31.35 for correct sequences, 36.87 for incorrect sequences), despite having better accuracy. The MiniCPM-2B-128k model demonstrates intermediate DTW distances. Across all models, correct sequences consistently show lower DTW distances than incorrect ones, validating DTW's effectiveness as a quality metric.

\subsubsection{Distributional Similarity}

\begin{table}[h]
\centering
\scalebox{0.9}{
\begin{tabular}{lcccccc}
\toprule
\textbf{Model} & \multicolumn{3}{c}{\textbf{SAE-based KL Divergence}} & \multicolumn{3}{c}{\textbf{Centroid-based KL Divergence}} \\\cmidrule{2-7}
& Corr-Orig & Incorr-Orig & Corr-Incorr & Corr-Orig & Incorr-Orig & Corr-Incorr \\
\midrule
MiniCPM-2B-128k & 0.178 & 0.391 & 0.236 & 0.086 & 0.258 & 0.235 \\
MiniCPM-1B-sft & 0.033 & 0.148 & 0.054 & 0.296 & 0.387 & 0.155 \\
MiniCPM-2B-sft & 0.068 & 0.157 & 0.011 & 0.357 & 0.375 & 0.300 \\
\bottomrule
\end{tabular}}
\caption{KL divergence measurements comparing distributional similarities between generated and original sequences. Measurements are shown for both direct representation (SAE) and centroid-based approaches. Lower values indicate better preservation of original sequence patterns. Corr-Orig: Correct vs Original, Incorr-Orig: Incorrect vs Original, Corr-Incorr: Correct vs Incorrect sequences.}
\label{tab:kl-divergence}
\end{table}
KL divergence measurements were conducted using both direct representations (SAE) and clustered centroids, as detailed in Table \ref{tab:kl-divergence}. SAE-based analysis shows the lowest KL divergence between correct and original sequences to be from the MiniCPM-1B-sft model (0.033), indicating superior preservation of the original distance distribution. MiniCPM-2B-128k exhibits the highest divergence (0.178), while MiniCPM-2B-sft maintains moderate divergence (0.068). Interestingly, the centroid-based analysis reveals different patterns, with MiniCPM-2B-128k showing the lowest KL divergence between correct and original sequences (0.086), while both fine-tuned models show higher centroid-based divergences. The distributional patterns (Figures~\ref{fig:sae_cos_sim_hist} and~\ref{fig:cent_cos_sim_hist}) demonstrate that incorrect generations tend to exhibit higher peak densities and narrower distributions, suggesting more repetitive patterns compared to both correct generations and original sequences.

\subsubsection{Entropy Analysis}

\begin{table}[h]
\centering
\begin{tabular}{lccc}
\toprule
\textbf{Model} & \textbf{Original} & \textbf{Correct} & \textbf{Incorrect} \\
\midrule
MiniCPM-2B-128k & 5.50 & 5.68 & 5.87 \\
MiniCPM-1B-sft & 5.41 & 5.42 & 5.58 \\
MiniCPM-2B-sft & 5.62 & 5.51 & 5.64 \\
\bottomrule
\end{tabular}
\caption{Entropy measurements quantifying sequence diversity across different models. Values are shown for original sequences and both correct and incorrect generations. Values closer to the original indicate better preservation of original sequence diversity patterns. }
\label{tab:entropy}
\end{table}
The entropy measurements assess sequence diversity, as presented in Table \ref{tab:entropy}. The MiniCPM-1B-sft model demonstrates the closest entropy match with the original sequences, while MiniCPM-2B-128k shows slightly higher entropy than the original, indicating potential over-diversification. The MiniCPM-2B-sft model shows balanced entropy levels close to the original distribution.

\subsection{Sequence Length Analysis}
The analysis of sequence lengths reveals important differences between correct and incorrect generations across models, as shown in Table \ref{tab:sequence-lengths}. The original sequences have an average length of 130.56 tokens, providing a baseline for comparison.
\begin{table}[h]
\centering
\begin{tabular}{lccc}
\toprule
\textbf{Model} & \textbf{Original Length} & \textbf{Correct Length} & \textbf{Incorrect Length} \\
\midrule
MiniCPM-2B-128k & 130.56 & 148.08 & 206.85 \\
MiniCPM-1B-sft & 130.56 & 132.77 & 177.83 \\
MiniCPM-1B-sft & 130.56 & 125.52 & 149.39 \\
\bottomrule
\end{tabular}
\caption{Comparison of average sequence lengths across different models and generation categories. The original sequence length serves as a reference point for evaluating generation fidelity. Notably, incorrect generations consistently produce longer sequences across all models, suggesting a correlation between sequence length deviation and generation quality.}
\label{tab:sequence-lengths}
\end{table}
The MiniCPM-1B-sft model demonstrates the closest length alignment for correct generations (132.77 tokens), with a 1.7\% deviation from the original length. The MiniCPM-2B-sft model slightly underestimates sequence lengths (125.52 tokens), while 2B-128k shows significant overestimation (148.08 tokens). Notably, incorrect generations consistently produce longer sequences across all models, with the most pronounced effect in MiniCPM-2B-128k (206.85 tokens, 58.4\% longer than the original). This pattern suggests that sequence length deviation might serve as an additional indicator of generation quality.

%% file: text/5_conclusion.tex
\section{Conclusion}

From our experiments, MiniCPM-2B-sft achieves the highest accuracy: it shows larger structural deviations (DTW distances), suggesting a potential trade-off between accuracy and structural preservation. This is most likely because adherence to the reference graph and creative reasoning traces that lead to high-quality generation and the correct answers couldn't coexist. The divergence between SAE and centroid-based measurements also highlights the importance of considering multiple representation levels (token-level vs. cluster-based) when evaluating sequence generation quality. 




These results show that exploitation-based methods alone are insufficient to guide generation quality. Instead, balancing both exploitation and exploration is crucial to guide model generation to search for the correct answer. Our proposed SAE-based technique is a scalable way to supervise intermediate token-level generation and balance exploitation and exploration through reward models.

Looking forward, there are many possible improvements to this framework that future research can resolve. The most immediate improvement is likely the design of a better reward function. We imagine that metrics that combine rewards calculated from the graph, cosine distance between consecutive tokens, and other multi-level structural and semantic evaluations can help us better understand and refine our proposed tradeoff between exploitation and exploration.

Another promising direction is to develop integrated metrics combining structural, distributional, and semantic aspects to provide a more comprehensive quality assessment. Particularly, investigating the relationship between distributional patterns and generation quality could lead to better early detection mechanisms for incorrect generations.

Given the consistent length and distributional deviations in incorrect generations, we can also look towards developing techniques to explicitly diverge from the "incorrect" pathway during generation in order to help improve output quality. The clear differences in distributional patterns between correct and incorrect generations also suggest the potential to develop better, more interpretable indicators.

Overall, we hope that these insights using our SAE-based clustering and graph-based technique can help lead the way toward more efficient and higher quality reasoning systems.

%% file: text/6_appendix.tex
\appendix
\section{Appendix}

\begin{figure}[!ht]
    \centering
    \begin{subfigure}{0.48\linewidth}
        \centering
        \includegraphics[width=\linewidth]{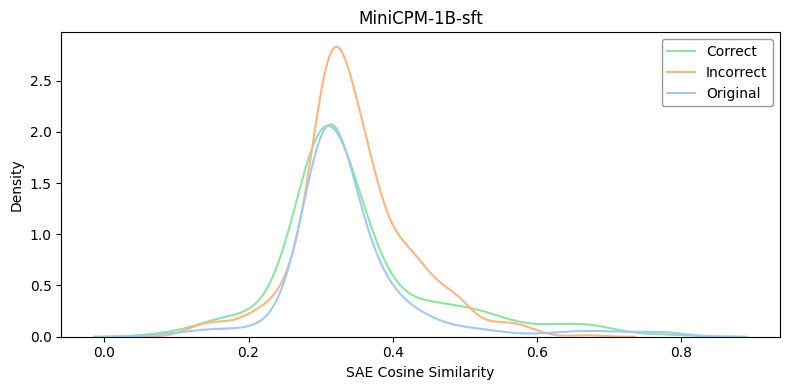}
        \caption{MiniCPM-1B-sft}
        \label{fig:sae_cos_sim_hist1}
    \end{subfigure}
    \hfill
    \begin{subfigure}{0.48\linewidth}
        \centering
        \includegraphics[width=\linewidth]{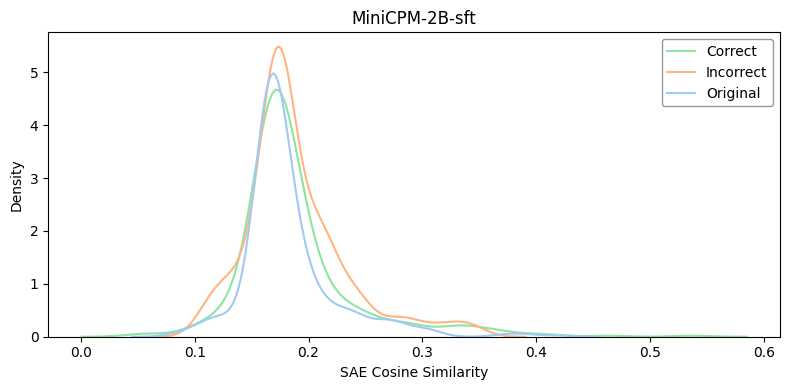}
        \caption{MiniCPM-2B-sft}
        \label{fig:sae_cos_sim_hist2}
    \end{subfigure}
    
    \vspace{1em}
    
    \begin{subfigure}{0.48\linewidth}
        \centering
        \includegraphics[width=\linewidth]{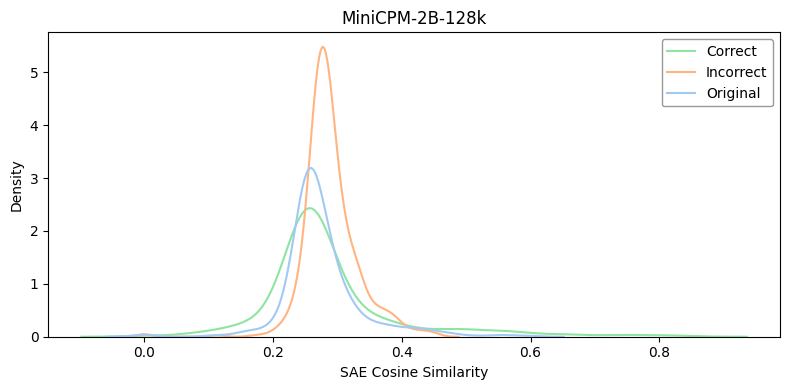}
        \caption{MiniCPM-2B-128k}
        \label{fig:sae_cos_sim_hist3}
    \end{subfigure}
    \caption{Distribution of SAE cosine similarities between consecutive sequence elements across different models. Each subplot compares distributions between correctly generated sequences (green), incorrectly generated sequences (orange), and original sequences (blue). (a) MiniCPM-1B-sft shows relatively lower peak density but better alignment between correct and original distribution, (b) MiniCPM-2B-sft demonstrates higher peak density with better distributional alignment, and (c) MiniCPM-2B-128k exhibits the highest peak density but shows notable deviation from the original distribution pattern. The curves represent kernel density estimations, revealing the underlying probability distribution of the cosine similarities. Higher similarity values indicate stronger semantic relationships between consecutive elements in the sequences.}
    \label{fig:sae_cos_sim_hist}
\end{figure}

\begin{figure}[!ht]
    \centering
    \begin{subfigure}{0.48\linewidth}
        \centering
        \includegraphics[width=\linewidth]{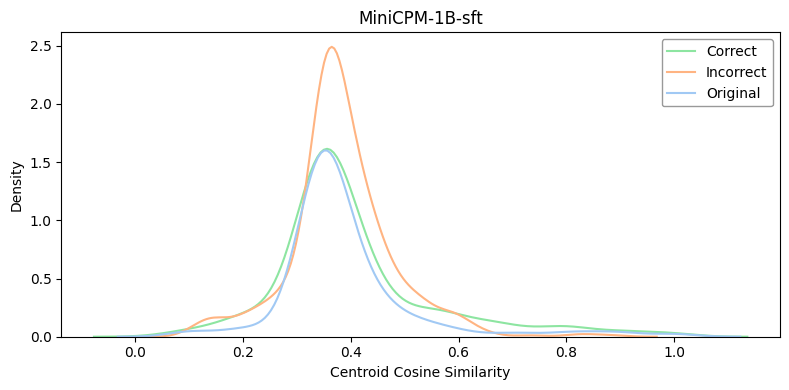}
        \caption{MiniCPM-1B-sft}
        \label{fig:cent_cos_sim_hist1}
    \end{subfigure}
    \hfill
    \begin{subfigure}{0.48\linewidth}
        \centering
        \includegraphics[width=\linewidth]{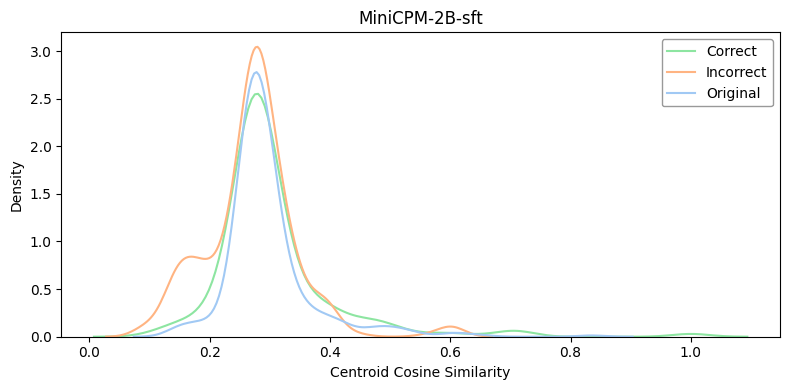}
        \caption{MiniCPM-2B-sft}
        \label{fig:cent_cos_sim_hist2}
    \end{subfigure}
    
    \vspace{1em}
    
    \begin{subfigure}{0.48\linewidth}
        \centering
        \includegraphics[width=\linewidth]{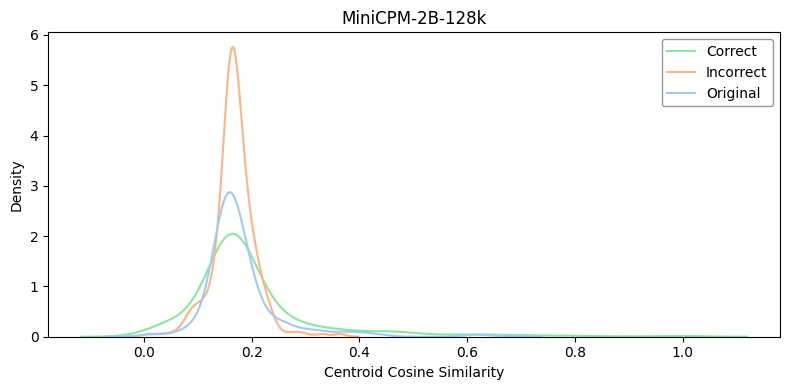}
        \caption{MiniCPM-2B-128k}
        \label{fig:cent_cos_sim_hist3}
    \end{subfigure}
    \caption{Distribution of centroid cosine similarities between consecutive sequence elements across different models. Each subplot compares distributions between correctly generated sequences (green), incorrectly generated sequences (orange), and original sequences (blue). (a) MiniCPM-1B-sft shows relatively lower peak density but better alignment between correct and original distribution, (b) MiniCPM-2B-sft demonstrates higher peak density with better distributional alignment, and (c) MiniCPM-2B-128k exhibits the highest peak density but shows notable deviation from the original distribution pattern. This centroid-based analysis complements the fine-grained SAE similarity distributions shown in Figure \ref{fig:sae_cos_sim_hist}.}
    \label{fig:cent_cos_sim_hist}
\end{figure}